\documentclass[11pt,a4paper]{article}
\usepackage[hyperref]{acl2020}
\usepackage{times}
\usepackage{latexsym}

\usepackage{microtype}

\aclfinalcopy 

\usepackage{url}
\usepackage{amsmath}
\usepackage{algorithm}
\usepackage{verbatim}
\usepackage{multirow}
\usepackage{multicol}

\usepackage{soul}

\usepackage[noend]{algpseudocode}

\usepackage{graphicx}  
\usepackage{float}

\title{DeSePtion: Dual Sequence Prediction and Adversarial Examples for Improved Fact-Checking}

\author{Christopher Hidey,\textsuperscript{1}\thanks{\hspace{.1cm}Work completed in part at Amazon} 
  \hspace{.1cm}Tuhin Chakrabarty,\textsuperscript{1} 
  Tariq Alhindi,\textsuperscript{1} 
  Siddharth Varia,\textsuperscript{1}\\
  \textbf{Kriste Krstovski,\textsuperscript{1,2}
  Mona Diab,\textsuperscript{3,4}\footnotemark[1] \hspace{.1cm}and
  Smaranda Muresan\textsuperscript{1,2}}\\
  \textsuperscript{1}Department of Computer Science, Columbia University \\
  \textsuperscript{2}Data Science Institute, Columbia University\\
  \textsuperscript{3}Facebook AI\\
  \textsuperscript{4}Department of Computer Science, George Washington University\\
  {\tt \{chidey, tariq, smara\}@cs.columbia.edu, mtdiab@email.gwu.edu}\\ 
  {\tt \{tuhin.chakrabarty, sv2504, kriste.krstovski\}@columbia.edu}}

\date{}

\begin{document}
\maketitle
\begin{abstract}

The increased focus on misinformation has spurred development of data and systems for detecting the veracity of a claim as well as retrieving authoritative evidence.
The Fact Extraction and VERification (FEVER) dataset provides such a resource for evaluating end-to-end fact-checking, requiring retrieval of evidence from Wikipedia to validate a veracity prediction.
We show that current systems for FEVER are vulnerable to three categories of realistic challenges for fact-checking -- multiple propositions, temporal reasoning, and ambiguity and lexical variation  --
and introduce a resource with these types of claims. Then we present a system designed to be resilient to these ``attacks'' using multiple pointer networks for document selection and jointly modeling a sequence of evidence sentences and veracity relation predictions. 
We find that in handling these attacks we obtain state-of-the-art results on FEVER, largely due to improved evidence retrieval. 

\end{abstract}

\section{Introduction}

The growing presence of biased, one-sided, and often altered discourse, is posing a challenge to our media platforms from newswire to social media \cite{Vosoughi1146}. To overcome this challenge, fact-checking has emerged as a necessary part of journalism, where experts examine "check-worthy" claims \cite{Hassan:2017:TAF:3097983.3098131} published by others for their ``shades'' of truth (e.g., FactCheck.org or PolitiFact). However, this process is time-consuming, and thus building computational models for automatic fact-checking has become an active area of research \cite{graves}. Advances were made possible by new open source datasets and shared tasks: the Fact Extraction and Verification Shared Task (FEVER) 1.0 and 2.0 \cite{N18-1074,thorne2019adversarial}, SemEval 2019 Shared Task 8: Fact-Checking in Community Forums \cite{mihaylova2019semeval},  and LIAR(+) datasets with claims from PolitiFact \cite{wang2017liar,alhindi-etal-2018-evidence}. 

\begin{figure}
\small
\centering
\begin{tabular}{|p{7cm}|}
\hline
\textbf{Claim:} Murda Beatz$'$s real name is Marshall Mathers. \\
\textbf{Evidence: [Murda Beatz]}  Shane Lee Lindstrom (born February 11,
1994), known professionally as Murda Beatz, is a
Canadian hip hop record producer and songwriter from
Fort Erie, Ontario. \\
\textbf{Label:} REFUTES \\
\hline
\end{tabular}
\caption{Example from FEVER 1.0 Dataset}
\label{figure:example}
\end{figure}

The FEVER 1.0 shared task dataset \cite{N18-1074} has enabled the development of end-to-end fact-checking systems, requiring document retrieval and evidence sentence extraction to corroborate a veracity relation prediction (supports, refutes, not enough info). 
An example is given in Figure \ref{figure:example}. 
Since the claims in FEVER 1.0 were manually written using information from Wikipedia, the dataset may lack linguistic challenges that occur in verifying naturally occurring check-worthy claims, such as temporal reasoning or lexical generalization/specification. 
\newcite{thorne2019adversarial} designed a second shared task (FEVER 2.0) for participants to create adversarial claims (``attacks'') to break state-of-the-art systems and then develop systems to resolve those attacks.

We present a \textbf{novel dataset of adversarial examples} for fact extraction and verification in three challenging categories: 1) multiple propositions (claims that require multi-hop document or sentence retrieval); 2) temporal reasoning (date comparisons, ordering of events); and 3) named entity ambiguity and lexical variation (Section \ref{sec:adversarial_claim_development}). We show that \textbf{state-of-the-art systems are vulnerable} to adversarial attacks from this dataset (Section \ref{sec:results}).
In addition, we take steps toward addressing these vulnerabilities, presenting a system for end-to-end fact-checking that brings \textbf{two novel contributions using pointer networks}: 1) a document ranking model; and 2) a joint model for evidence sentence selection and veracity relation prediction framed as a sequence labeling task  (Section \ref{sec:methods}).
 Our new system achieves \textbf{state-of-the-art results for FEVER} and we present an \textbf{evaluation of our models} including ablation studies (Section \ref{sec:results}). Data and code will be released to the  community.\footnote{\url{https://github.com/chridey/fever2-columbia}}

\section{Related Work}
Approaches for predicting the veracity of naturally-occurring claims have focused on statements fact-checked by journalists or organizations such as  PolitiFact.org \cite{W14-2508,alhindi-etal-2018-evidence}, news articles \cite{raofakenews}, or answers in community forums \cite{DBLP:journals/corr/abs-1803-03178, mihaylova2019semeval}. However, those datasets are not suited for end-to-end fact-checking as they provide sources and evidence while FEVER \cite{N18-1074} requires retrieval.

Initial work on FEVER focused on a pipeline approach of retrieving documents, selecting sentences, and then using an entailment module \cite{malon-2018-team, hanselowski-etal-2018-ukp,fever1}; the winning entry for the FEVER 1.0 shared task \cite{fever1.0} used three homogeneous neural  models. Other work has jointly learned either evidence extraction and question answering \cite{fever2} or sentence selection and relation prediction \cite{twowingos, hidey-diab-2018-team}; unlike these approaches, we use the same sequential evidence prediction architecture for both document and sentence selection, jointly predicting a sequence of labels in the latter step. More recently, \newcite{fever3} proposed a graph-based framework for multi-hop retrieval, whereas we model evidence sequentially.

Language-based adversarial attacks have often involved transformations of the input such as phrase insertion to distract question answering systems \cite{jia-liang-2017-adversarial} or to force a model to always make the same prediction \cite{wallace-etal-2019-universal}. Other research has resulted in adversarial methods for paraphrasing with universal replacement rules \cite{adver1} or lexical substitution \cite{DBLP:journals/corr/abs-1804-07998, ren-etal-2019-generating}. 
While our strategies include insertion and replacement, we focus specifically on challenges in fact-checking. The task of natural language inference \cite{D15-1075, N18-1101} provides similar challenges: examples for numerical reasoning and lexical inference have been shown to be difficult \cite{glockner2018breaking, nie2019adversarial} and improved models on these types are likely to be useful for fact-checking.
Finally, \cite{thorne2019adversarial} provided a baseline for the FEVER 2.0 shared task with entailment-based perturbations. Other participants generated adversarial claims using implicative phrases such as ``not clear'' \cite{kim-allan-2019-fever} or GPT-2 \cite{niewinski-etal-2019-gem}.
In comparison, we present a diverse set of  attacks motivated by realistic, challenging categories and further develop models to address those attacks.

\section{Problem Formulation and Datasets}
\label{sec:data}

We address the end-to-end fact-checking problem in the context of FEVER \cite{N18-1074}, a task where a system is required to verify a claim by providing evidence from Wikipedia. To be successful, a system needs to predict both the correct veracity relation-- supported ({\sc s}), refuted ({\sc r}), or not enough information ({\sc nei})-- and the correct set of evidence sentences (not applicable for {\sc nei}). The \textbf{FEVER 1.0} dataset \cite{N18-1074} was created by extracting sentences from popular Wikipedia pages and mutating them with paraphrases or other edit operations to create a claim.  Then, each claim was labeled and paired with evidence or the empty set for NEI.  Overall, there are 185,445 claims, of which 90,367 are {\sc s}, 40,107 are {\sc r}, and 45,971 are NEI. 
\newcite{thorne2019adversarial} introduced an adversarial set up for the \textbf{FEVER 2.0} shared task -- participants submitted claims to break existing systems and a system designed to withstand such attacks. The organizers provided a baseline of 1000 adversarial examples with negation and entailment-preserving/-altering transformations and this set was combined with examples from participants to form the FEVER 2.0 dataset.
Table \ref{table:data_stats} shows the partition of FEVER 1.0 and 2.0 data (hereafter FV1/FV2-train/dev/test).

\begin{table}[h!]
\small
    \centering
    \begin{tabular}{|l|r|r|r|}
    \hline
    Dataset & Train & Dev. & Blind Test  \\
    \hline
    FEVER 1.0     & 145,449 & 19,998 & 19,998\\
    \hline
    FEVER 2.0     & -- & 1,174 & 1,180 \\
    \hline
    \end{tabular}
    \caption{FEVER Dataset Statistics}
    \label{table:data_stats}
\end{table}

\section{Adversarial Dataset for Fact-checking} \label{sec:adversarial_claim_development}
While the FEVER dataset is a valuable resource, our goal is to evaluate complex adversarial claims which resemble check-worthy claims found in news articles, speeches, debates, and online discussions. We thus propose three types of  attacks based on analysis of FV1 or prior literature: those using multiple propositions, requiring temporal and numerical reasoning, and involving lexical variation. 

For the \textbf{multi-propositional} type, \newcite{graves} notes that
professional fact-checking organizations need to synthesise evidence from multiple sources; automated systems struggle with claims such as \emph{``Lesotho is the smallest country in Africa.''}  In FV1-dev, 83.18\% of S and R claims require only a single piece of evidence and 89\% require only a single Wikipedia page.
Furthermore, our previous work on FEVER 1.0 found that our model can fully retrieve 86\% of evidence sentences from Wikipedia when only a single sentence is required, but the number drops to 17\% when 2 sentences are required and 3\% when 3 or more sentences are required \cite{hidey-diab-2018-team}.

For the second type, check-worthy claims are often numerical \cite{full_fact} and \textbf{temporal reasoning} is especially challenging  \cite{DBLP:conf/coling/MirzaT16}.
\newcite{rashkin-etal-2017-truth} and \newcite{Jiang:2018:LSU:3290265.3274351} showed that numbers and comparatives are indicative of truthful statements in news, but the presence of a date alone does not indicate its veracity. In FV1-dev, only 17.81\% of the claims contain dates and 0.22\% contain time information.\footnote{As determined by NER using Spacy: https://spacy.io}
To understand how current systems perform on these types of claims, we evaluated three state-of-the-art systems from FEVER 1.0  \cite{hanselowski-etal-2018-ukp, yoneda-etal-2018-ucl, fever1.0}, and examined the predictions where the systems disagreed.  We found that in characterizing these predictions according to the named entities present in the claims, the most frequent types were numerical and temporal (such as percent, money, quantity, and date).

Finally, adversarial attacks for \textbf{lexical variation}, where words may be inserted or replaced or changed with some other edit operation, have been shown to be effective for similar tasks such as natural language inference \cite{nie2019adversarial} and question answering \cite{jia-liang-2017-adversarial}, so we include these types of attacks as well.
For the fact-checking task, models must match words and entities across claim and evidence to make a veracity prediction. As claims often contain ambiguous entities \cite{C18-1283} or lexical features indicative of credibility \cite{nakashole-mitchell-2014-language}, we desire models resilient to minor changes in entities \cite{hanselowski-etal-2018-ukp} and words \cite{DBLP:journals/corr/abs-1804-07998}.

We thus create an adversarial dataset  of 1000 examples, with 417 multi-propositional, 313 temporal and 270 lexically variational. Representative examples are provided in Appendix \ref{appendix:attack_examples}.

\paragraph{Multiple Propositions}
Check-worthy claims often consist of multiple propositions \cite{graves}. 
In the FEVER task, checking these claims may require retrieving evidence sequentially
after resolving entities and events, understanding 
discourse connectives, and evaluating each proposition.
  
Consider the claim \emph{``Janet Leigh was from New York and was an author.''}
The Wikipedia page \textbf{[Janet Leigh]} contains evidence that she was an author, but makes no mention of New York.
We generate new claims of the {\sc conjunction} type \emph{automatically} by mining claims from FV1-dev and extracting entities from the subject position. We then combine two claims by replacing the subject in one sentence with a discourse connective such as ``and.''
The new label is S if both original claims are S, R if at least one claim is R, and NEI otherwise.

While {\sc conjunction} claims provide a way to evaluate multiple propositions about a single entity, these claims only require evidence from a single page; hence we create new examples requiring reasoning over multiple pages. 
To create {\sc multi-hop} examples,
we select claims from FV1-dev whose evidence obtained from a single page \textit{P} contains at least one other entity having a valid page \textit{Q}. We then modify the claim by appending information about the entity which can be verified from \textit{Q}. 
For example, given the claim \emph{``The Nice Guys is a 2016 action comedy film.''} we make a multi-hop claim by obtaining the page  \textbf{[Shane Black]} (the director) and appending the phrase \emph{``directed by a Danish screenwriter known for the film Lethal Weapon.``}

While multi-hop retrieval provides a way to evaluate the {\sc s} and {\sc r} cases, composition of multiple propositions may also be necessary for NEI, as the relation of the claim and evidence may be changed by more general/specific phrases.
We thus add {\sc additional unverifiable propositions} that change the gold label to NEI. We selected claims from FV1-dev and added propositions which have no evidence in Wikipedia (e.g. for the claim \emph{``Duff McKagan is an American citizen,''} we can add the reduced relative clause \textit{``born in Seattle``}).

\paragraph{Temporal Reasoning}
Many check-worthy claims contain dates or time periods and to verify them requires models that handle temporal reasoning \cite{thorne-vlachos-2017-extensible}.

In order to evaluate the ability of current systems to handle temporal reasoning we modify claims from  FV1-dev. 
More specifically, using claims with the phrase "in $<$date$>$" we \emph{automatically} generate seven modified claims using simple {\sc date manipulation} heuristics: 
arithmetic (e.g., \emph{``in 2001"} $\rightarrow$ \emph{``4 years before 2005"}), range (\emph{``in 2001"} $\rightarrow$ \emph{``before 2008"}), and verbalization (\emph{``in 2001"} $\rightarrow$ \emph{``in the first decade of the 21st century"}). 
 
We also create examples requiring {\sc multi-hop temporal reasoning}, where the system must evaluate an event in relation to another. 
Consider the {\sc s} claim \emph{``The first governor of the Indiana Territory lived long enough to see it become a state.''} A system must resolve entity references (Indiana Territory and its first governor, William Henry Harrison) and compare dates of events (the admittance of Indiana in 1816 and death of Harrison in 1841).
While multi-hop retrieval may resolve references, the model must understand the meaning of \emph{``lived long enough to see''} and evaluate the comparative statement.
To create claims of this type, we mine Wikipedia by  selecting a page $X$ and extracting sentences with the pattern ``is/was/named the $A$ of $Y$'' (e.g. $A$ is \emph{``first governor''}) where $Y$ links to another page.  Then we manually create temporal claims by examining dates on $X$ and $Y$ and describing the relation between the entities and events.

\paragraph{Named Entity Ambiguity and Lexical Variation}
As fact-checking systems are sensitive to lexical choice \cite{nakashole-mitchell-2014-language, rashkin-etal-2017-truth}, we consider how  variations in entities and words may affect veracity relation prediction.

{\sc Entity disambiguation} has been shown to be important for retrieving the correct page for an entity among multiple candidates \cite{hanselowski-etal-2018-ukp}.
To create examples that contain ambiguous entities we selected claims from FV1-dev where at least one Wikipedia disambiguation page was returned by the Wikipedia python API.\footnote{https://pypi.org/project/wikipedia/} We then created a new claim using one of the documents returned from the disambiguation list. For example the claim \textit{``Patrick Stewart is someone who does acting for a living.''} returns a disambiguation page, which in turn gives a list of pages such as \textbf{[Patrick Stewart]} and \textbf{[Patrick Maxwell Stewart]}.

Finally, as previous work has shown that neural models are vulnerable to {\sc lexical substitution} \cite{DBLP:journals/corr/abs-1804-07998}, we apply their genetic algorithm approach to replace words via counter-fitted embeddings. We make a claim adversarial to a model fine-tuned on claims and gold evidence by replacing synonyms, hypernyms, or hyponyms, e.g. \textit{created} $\rightarrow$ \textit{established}, \textit{leader} $\rightarrow$ \textit{chief}. We manually remove ungrammatical claims or incorrect relations.

\section{Methods}
\label{sec:methods}

\begin{figure}[h]
\includegraphics[scale=0.55, trim={8.5cm 3cm 1cm 3cm}, clip]{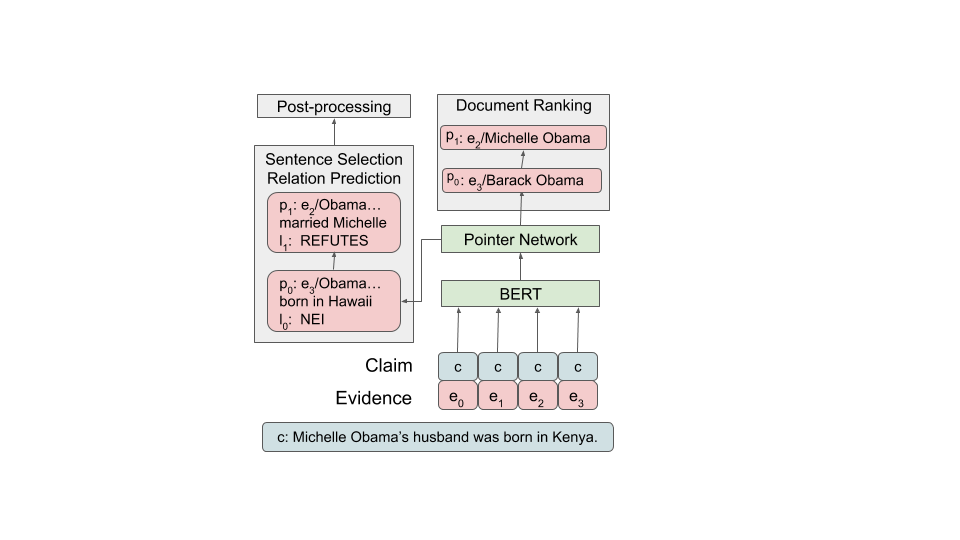}
\caption{\label{figure:joint_ptr} Pointer network architecture. Claim and evidence (page title or sentence) are embedded with BERT and evidence is sequentially predicted (for sentence selection the relation sequence is jointly predicted).}
\end{figure}

Verifying check-worthy claims such as those in Section \ref{sec:adversarial_claim_development} requires a system to 1) make sequential decisions to handle multiple propositions, 2) support temporal reasoning, and 3) handle ambiguity and complex lexical relations. To address the first requirement we make use of a pointer network \cite{NIPS2015_5866} in two novel ways: i) to re-rank candidate documents and ii) to jointly predict a sequence of evidence sentences and veracity relations in order to compose evidence (Figure \ref{figure:joint_ptr}). To address the second we add a post-processing step for simple temporal reasoning. To address the third we use rich, contextualized representations. Specifically, we fine-tune BERT \cite{bert} as this model has shown excellent performance on related tasks and was pre-trained on Wikipedia. 
\begin{figure}[h]
\includegraphics[ scale=0.45, trim={8cm 1cm 0 5cm}, clip]{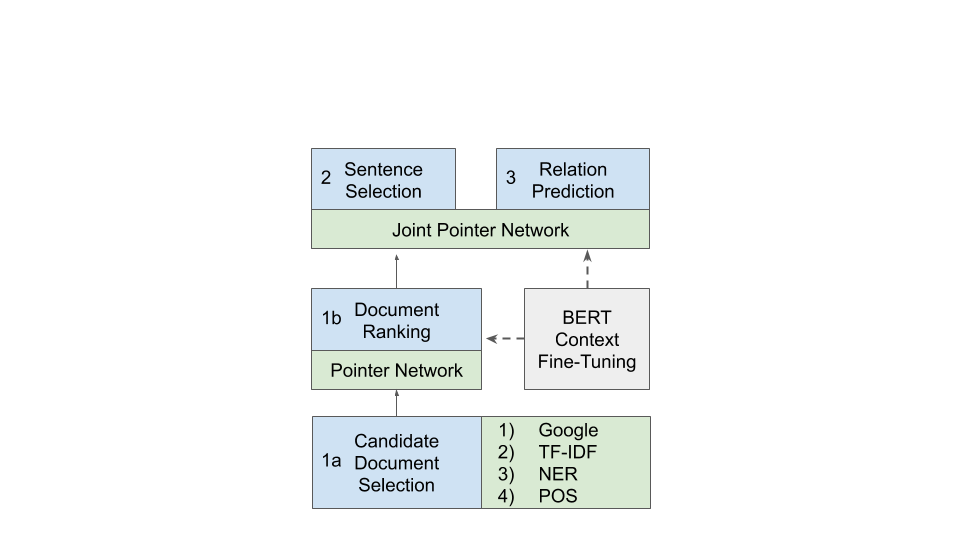}
\caption{Our FEVER pipeline: 1) Retrieving Wikipedia pages by selecting an initial candidate set (1a) and ranking the top $D$ (1b); 2) Identifying the top $N$ sentences; 3) Predicting supports, refutes, or not enough info. Dashed arrows indicate fine-tuning steps.}
\label{figure:pipeline}
\end{figure}

Our full pipeline is presented in Figure \ref{figure:pipeline}.  We first identify an initial \textbf{candidate set of documents} (1a) by combining the top $M$ pages from a TF-IDF search using DrQA \cite{P17-1171} with pages from the approach of \newcite{W18-5521}, which provides results from Google search and predicted named entities and noun phrases.  Then, we perform \textbf{document ranking} by selecting the top $D<M$ pages with a pointer network (1b). Next, an $N$-long sequence of evidence sentences (2) and veracity relation labels (3) are \textbf{predicted jointly by another pointer network}. 

Prior to training, we fine-tune BERT for document and sentence ranking on claim/title and claim/sentence pairs, respectively. Each claim and evidence pair in the FEVER 1.0 dataset has both the title of the Wikipedia article and at least one sentence associated with the evidence, so we can train on each of these pairs directly. For the claim \textit{``Michelle Obama's husband was born in Kenya''}, shown in Figure \ref{figure:joint_ptr}, we obtain representations by pairing this claim with evidence sentences such as \textit{``Obama was born in Hawaii''} and article titles such as \textbf{[Barack Obama]}.

The core component of our approach is the pointer network, as seen in Figure \ref{figure:joint_ptr}.
Unlike our previous work \cite{hidey-diab-2018-team}, we use the pointer network to re-rank candidate documents and jointly predict a sequence of evidence sentences and relations.  Given a candidate set of evidence (as either document titles or  sentences) and a respective fine-tuned BERT model, we extract features for every claim $c$ and evidence $e_p$ pair by summing the $[CLS]$ embedding for the top 4 layers (as recommended by \newcite{bert}): 
\begin{equation}
m_p = BERT(c, e_p)
\label{equation:bert}
\end{equation}

Next, to select the top $k$ evidence, we use a pointer network over the  evidence for claim $c$ to extract evidence recurrently by computing the extraction probability $P(p_t|p_0 \cdots p_{t-1})$ for evidence $e_p$ at time $t<k$. At time $t$, we update the hidden state $z_t$ of the pointer network decoder.  Then we compute the weighted average $h_t^q$ of the entire evidence set using $q$ hops over the evidence \cite{glimpse, NIPS2015_5846}:\footnote{Initially, $h^{t,0}$ is set to $z_t$. $v_h$, $W_{g}$, and $W_{a}$ are learned.}
\begin{equation}
\begin{split}
& \alpha_t^o = \text{softmax}(v^T_h \tanh(W_{g}m_p + W_{a} h_t^{o-1})) \\
& h_t^o = \sum_j \alpha_t^o W_{g} m_j
\end{split}
\label{equation:equation_hidden_state}
\end{equation}
We concatenate $m_p$ and $h_t^q$ and use a multi-layer perceptron (MLP) to predict $p_t$.  The loss is then:
\begin{equation}
\mathcal{L}(\theta_{ptr}) = -1/k \sum_{t=0}^{k-1} \log P_{\theta_{ptr}} (p_t | p_{0:t-1})
\label{equation:equation_gold_ptr_loss}
\end{equation}

We train on gold evidence and perform inference with beam search for both document ranking (Section \ref{subsec:document_ranking}) and joint sentence selection and relation prediction (Section \ref{subsec:joint_pointer}).

\subsection{Document Ranking} 
\label{subsec:document_ranking}
In order to obtain representations as input to the pointer network for document ranking, we leverage the fact that Wikipedia articles all have a title (e.g. \textbf{[Barack Obama]}), and fine-tune BERT  on title and claim pairs, in lieu of examining the entire document text (which due to its length is not suitable for BERT). Because the title often overlaps lexically with the claim (e.g. \textbf{[Michelle Obama]}), we can train the model to locate the title in the claim.
Furthermore, the words in the title co-occur with words in the article (e.g. \textit{Barack} and \textit{Michelle}), which the pre-trained BERT language model may be attuned to.
We thus fine-tune a classifier on a dataset created from title and claim pairs (where positive examples are titles of gold evidence pages and negative are randomly sampled from our candidate set), obtaining 90.0\% accuracy.
Given the fine-tuned model, we extract features using Equation \ref{equation:bert} where $e_p$ is a title, and use Equation \ref{equation:equation_gold_ptr_loss} to learn to predict a sequence of titles as in Figure \ref{figure:joint_ptr}.

\subsection{Joint Sentence Selection and Relation Prediction}
\label{subsec:joint_pointer}
The sentence selection and relation prediction tasks are closely linked, as predicting the correct evidence is necessary for predicting {\sc s} or {\sc r} and the representation should reflect the interaction between a claim and an evidence set. Conversely, if a claim and an evidence set are unrelated, the model should predict {\sc nei}. We thus jointly model this interaction by sharing the parameters of the pointer network - the hidden state of the decoder is used for both tasks and the models differ only by a final MLP.

\paragraph{Sentence Selection}
Similar to our document selection fine-tuning approach, we fine-tune a classifier on claim and evidence sentence pairs to obtain BERT embeddings.  However, instead of training a binary classifier for the presence of valid evidence
we train directly on veracity relation prediction, which is better suited for the end task. 
We create a dataset by pairing each claim with its set of gold evidence sentences. 
As gold evidence is not available for {\sc nei} relations, we sample sentences from our candidate documents to maintain a balanced dataset.
We then fine-tune a BERT classifier on relation prediction, obtaining 93\% accuracy.
Given the fine-tuned model, we extract features using Equation \ref{equation:bert} where $e_p$ is a sentence, and use Equation \ref{equation:equation_gold_ptr_loss} to learn to predict a sequence of sentences.

\paragraph{Relation Prediction}
In order to closely link relation prediction with evidence prediction, we re-frame the task as a sequence labeling task. In other words, rather than make a single prediction given all evidence sentences, we make one prediction at every timestep during decoding to model the relation between the claim and \emph{all evidence retrieved to that point}.
This approach provides three benefits: it allows the model to better handle noise (when an incorrect evidence sentence is predicted), to handle multi-hop inference (to model the occurrence of switching from {\sc nei} to {\sc s}/{\sc r}), and to effectively provide more training data (for $k=5$ timesteps we have 5 times as many relation labels). 
For the claim in Figure \ref{figure:joint_ptr}, the initial label sequence is {\sc nei} and {\sc r} because the first evidence sentence by itself (the fact that Barack Obama was born in Hawaii) would not refute the claim. Furthermore for $k=5$, the remaining sequence would be {\sc r}, {\sc r}, {\sc r}, as additional evidence (guaranteed to be non-contradictory in FEVER) would not change the prediction. On the other hand, given a claim that requires only a single piece of evidence, such as that in Figure \ref{figure:example}, the sequence would be {\sc r}, {\sc r}, {\sc r}, {\sc r}, {\sc r} if the correct evidence sentence was selected at the first timestep, {\sc nei}, {\sc r}, {\sc r}, {\sc r}, {\sc r} if the correct evidence sentence was selected at the second timestep, and so forth.

We augment the evidence sentence selection described previously to use the hidden state of the pointer network after $q$ hops (Equation \ref{equation:equation_hidden_state}) and an MLP to also predict a label 
at that time step, closely linking evidence and label prediction:
\begin{equation}
P(l_t) = \text{softmax}(W_{l2} \text{tanh}(W_{l1} h^o_t))
\label{equation:equation_label_prob}
\end{equation}
As with evidence prediction (Equation \ref{equation:equation_gold_ptr_loss}), when the gold label sequence is available, the loss term is:
\begin{equation}
\mathcal{L}(\theta_{rel\_seq}) = -1/k \sum_{t=0}^{k-1} \log P_{\theta_{rel\_seq}} (l_t)
\label{equation:equation_label_loss}
\end{equation}
When training, at the current timestep we use both the gold evidence, i.e. ``teacher forcing'' \cite{Williams:1989:LAC:1351124.1351135}, and the model prediction from the previous step, so that we have training data for NEI.
Combining Equations \ref{equation:equation_gold_ptr_loss} and \ref{equation:equation_label_loss}, our loss is:
\begin{equation}
\mathcal{L}(\theta) = \lambda \mathcal{L}(\theta_{ptr}) +  \mathcal{L}(\theta_{rel\_seq})
\label{equation:full_training_loss}
\end{equation}

Finally, to predict a relation at inference, we ensemble the sequence of predicted labels by averaging the probabilities over every time step.\footnote{The subset of timesteps was determined empirically: while at the final timestep the model is likely to have seen the correct evidence it also contains more noise; in future work we will experiment with alternatives.} 

\paragraph{Post-processing for Simple Temporal Reasoning} 

As neural models are unreliable for handling numerical statements, we introduce a  rule-based step to extract and reason about dates.
We use the Open Information Extraction  system of \newcite{stanovsky2018supervised} to extract tuples. 
For example, given the claim \emph{``The Latvian Soviet Socialist Republic was a republic of the Soviet Union 3 years after 2009,''} the system would identify \textbf{ARG0} as preceding the verb \emph{was} and \textbf{ARG1} following.
After identifying tuples in claims and predicted sentences, we discard those lacking dates (e.g. \textbf{ARG0}).  Given more than one candidate sentence, we select the one 
ranked higher by the pointer network. Once we have both the claim and evidence date-tuple we apply one of three rules to resolve the relation prediction based on the corresponding temporal phrase.  We either evaluate whether the evidence date is between two dates in the claim (e.g. \emph{between/during/in}), we add/subtract $x$ years from the date in the claim and compare to the evidence date (e.g. \emph{x years/days before/after}), or compare the claim date directly to the evidence date (e.g. \emph{before/after/in}). For the date expression \emph{``3 years after 2009,''} we compare the year \emph{2012} to the date in the retrieved evidence (\emph{1991}, the year the USSR dissolved)
and label the claim as {\sc r}.

\section{Experiments and Results}
\label{sec:results}

We evaluate our dataset and system as part of the FEVER 2.0 shared task in order to validate the vulnerabilities introduced by our adversarial claims (Section \ref{sec:adversarial_claim_development}) and the solutions proposed by our system (Section \ref{sec:methods}).
We train our system on FV1-train and evaluate on FV1/FV2-dev/test (Section \ref{sec:data}). We report  \textit{accuracy} (percentage of correct labels) and \textit{recall} (whether the gold evidence is contained in selected evidence at $k=5$). We also report the \textit{FEVER score},  the percentage of correct evidence sentences (for {\sc s} and {\sc r}) that also have correct labels, and \textit{potency}, the inverse FEVER score (subtracted from one) for evaluating adversarial claims. 

{\bf Our Baseline-RL:} For baseline experiments, to compare different loss functions, we use the approach of \newcite{W18-5521} for document selection and ranking, the reinforcement learning (RL) method of \newcite{P18-1063} for sentence selection, and BERT \cite{bert} for relation prediction. 
The RL approach using a pointer network is detailed by \newcite{P18-1063} for extractive summarization, with the only difference that we use our fine-tuned BERT on claim/gold sentence pairs to represent each evidence sentence in the pointer network (as with our full system) and use the FEVER score as a reward. The reward is obtained by selecting sentences with the pointer network and then predicting the relation using an MLP (updated during training) and the concatenation of all claim/predicted sentence representations with their maximum/minimum pooling.    

Hyper-parameters and settings for all experiments are detailed in Appendix \ref{appendix:hyps}.

\subsection{Adversarial Dataset Evaluation}
We present the performance of our adversarial claims, obtained by submitting to the shared task server.  We compare our claims to other participants in the FEVER 2.0 shared task (Table \ref{table:breakers_results}) and divided by attack type (Table \ref{table:breakers_data}). \textit{Potency} was macro-averaged across different fact-checking systems \cite{thorne2019adversarial}, correctness of labels was verified by shared task annotators,  and adjusted potency was calculated by the organizers as the potency of correct examples. Compared to other participants (Table \ref{table:breakers_results}), we presented a larger set of claims (501 in dev and 499 in test). 
We rank second in adjusted potency, but we provided a more diverse set than those created by the organizers or other participants. The organizers \cite{thorne2019adversarial} created adversarial claims using simple pattern-matching and replacement, e.g. quantifiers and negation. \newcite{niewinski-etal-2019-gem} trained a GPT-2-based model on the FEVER data and manually filtered disfluent claims. \newcite{kim-allan-2019-fever} considered a variety of approaches, the majority of which required understanding area comparisons between different regions or understanding implications (e.g. that ``not clear'' implies {\sc nei}).
While GPT-2 is effective, our approach is controllable and targeted at real-world challenges. Finally, Table \ref{table:breakers_data} shows that when we select our top 200 most effective examples (multi-hop reasoning and multi-hop temporal reasoning) and compare to the approaches of 
\newcite{niewinski-etal-2019-gem} and \newcite{kim-allan-2019-fever} (who both provided less than 204 examples total)
our potency is much higher.
In particular, multi-hop reasoning has a potency of 88\% for {\sc support} relations and 93\% for {\sc refutes} relations and multi-hop temporal reasoning obtains 98\% for {\sc support} and {\sc refutes} relations.

\begin{table}[h]
\small
\centering
\begin{tabular}{|l|l|l|l|l|}
\hline
Team & \# & Pot. & Corr. & Adj. \\
\hline
Organizer Baseline & 498 & 60.34 & 82.33 & 49.68 \\
\hline
\newcite{kim-allan-2019-fever} & 102 & 79.66 & 64.71 & 51.54 \\
\hline
\color{gray}\textbf{Ours} & \textbf{501} & \color{gray}\textbf{68.51} & \color{gray}\textbf{81.44} & \color{gray}\textbf{55.79} \\
\hline
\newcite{niewinski-etal-2019-gem} & 79 & 79.97 & 84.81 & 66.83 \\
\hline
\end{tabular}
\caption{The evaluation of our claims relative to other participants. \textbf{\#:} Examples in blind test \textbf{Pot:} Potency score \textbf{Corr.:} Percent grammatical and coherent with correct label and evidence \textbf{Adj.:} Adjusted potency}
\label{table:breakers_results}
\end{table}

\begin{table}[h]
\centering
\small
\begin{tabular}{|l|p{.55cm}|p{1.1cm}|p{1.1cm}|p{1.1cm}|}
\hline
Attack & M/A & \#S/P & \#R/P & \#NEI/P \\
\hline
\hline
Conjunct. & A & -/-  & 54/55\% & 75/63\%\\
Multi-hop & M & 100/88\%  & 88/93\% & 99/50\% \\
Add. Unver. & M & -/- & -/- & 50/50\%\\
\hline
\hline
Date Man. & A & 49/59\% & 129/80\% & 80/46\%\\
Mul. Temp. & M & 46/98\% & 5/98\% & 4/29\%\\
\hline
\hline
Entity Dis. & M & 46/50\% & -/- & -/- \\
Lexical Sub. & A* & 92/70\% & 57/70\%  & 25/38\%\\
\hline
\end{tabular}
\caption{\textbf{Attack:} Type of attack as described in Section \ref{sec:adversarial_claim_development}. \textbf{M/A:} Whether claims are generated manually (M), automatically (A), or verified manually (A*) \textbf{\#S:} Support examples  \textbf{\#R:} Refute examples \textbf{\#NEI} Not enough info examples \textbf{P:} Potency on Shared Task systems} 
\label{table:breakers_data}
\end{table}

\subsection{Evaluation against State-of-the-art}
In Tables \ref{table:fixers_fever1_blind_results} and \ref{table:breakers_blind_test_results} we compare Our System (Section \ref{sec:methods}) to recent work from teams that submitted to the shared task server for FEVER 1.0 and 2.0, respectively, including the results of Our Baseline-RL  system in Table \ref{table:breakers_blind_test_results}. 
Our dual pointer network approach obtains state-of-the-art results on the FEVER 1.0 blind test set (Table \ref{table:fixers_fever1_blind_results}) on all measures even over systems designed specifically for evidence retrieval \cite{fever2, fever3}, largely due to a notable improvement in recall (more than 3 points over the next system \cite{hanselowski-etal-2018-ukp}). We also find improvements in accuracy over the remaining pipeline systems, suggesting that joint learning helps. Compared to Our Baseline-RL,  Our System has 1.8 point improvement in FEVER score on FV1-test with 4 points on FV2-test. 
Notably, our system finishes second (with a score of 36.61) on the FEVER 2.0 shared task test set, even though our claims were designed to be challenging for our model.  The model of \newcite{malon-2018-team} performs especially well; they use a transformer-based architecture without pre-training but focus only on single-hop claims.

\begin{table}[h]
\center
\small
\begin{tabular}{|p{3.25cm}|p{.8cm}|p{.75cm}|l|}
\hline
System      & Acc. & Rec. & FEVER  \\ \hline
\newcite{hanselowski-etal-2018-ukp}       &65.46 & 85.19 & 61.58                 \\ \hline
\newcite{fever2}    &69.30 &76.30 & 61.80                 \\ \hline
\newcite{yoneda-etal-2018-ucl}       &67.62 & 82.84 & 62.52                \\ \hline
\newcite{fever1.0}   &68.16 & 71.51 & 64.21                  \\ \hline
\newcite{fever1}      &69.98 & 77.28 & 66.72               \\ \hline
\newcite{fever3}        & 71.60 & - & 67.10                \\ \hline
{\bf Our System}       & \textbf{72.47} & \textbf{88.39} & \textbf{68.80}                \\ \hline
\end{tabular}
\caption{Comparison with state of the art on FV1-test}  
    \label{table:fixers_fever1_blind_results}
\end{table}

\begin{table}[h]
\small
\centering
\begin{tabular}{|l|l|l|}
\hline
Team & FV1-test & FV2-test \\
\hline
\newcite{hanselowski-etal-2018-ukp}  & 61.58 & 25.35\\
\hline
\newcite{fever1.0}  & 64.21 & 30.47\\
\hline
\color{gray}\textbf{Our Baseline-RL}  & \color{gray}\textbf{67.08} & \color{gray}\textbf{32.92}\color{black}\\
\hline
\newcite{stammbach-neumann-2019-team}  & 68.46 & 35.82\\
\hline
\newcite{yoneda-etal-2018-ucl}		& 62.52 & 35.83\\
\hline
\textbf{Our System}  & \textbf{68.80} & 36.61\\
\hline
\newcite{malon-2018-team}  & 57.36 & \textbf{37.31}\\
\hline
\end{tabular}
\caption{Comparison of FEVER score to other shared-task systems (ordered by FV2-test FEVER score)}
\label{table:breakers_blind_test_results}
\end{table}

\subsection{System Component Ablation}
To better understand the improved performance of our system,  we present two ablation studies in Tables \ref{table:fever1_results} and \ref{table:fixers_fever2_results} on FV1 and FV2 dev, respectively.\footnote{Our system is significantly better on all metrics ($p < 0.001$ by the approximate randomization test).}

Table \ref{table:fever1_results} presents the effect of using different objective functions for sentence selection and relation prediction, compared to joint sentence selection and relation prediction in our full model. We compare Our System to Our Baseline-RL system as well as another baseline (\textit{Ptr}).
The \textit{Ptr} system is the same as Our Baseline-RL, except the pointer network and MLP are not jointly trained with RL but independently using gold evidence and predicted evidence and relations, respectively.
Finally, the Oracle upper bound presents the maximum possible recall after our document ranking stage, compared to 94.4\% for \newcite{W18-5521}, and relation accuracy (given the MLP trained on 5 sentences guaranteed to contain gold evidence). We find that by incorporating the relation sequence loss, we improve the evidence recall significantly 
relative to the oracle upper-bound, reducing the relative error by 50\% while also obtaining improvements on relation prediction, even over a strong RL baseline.  Overall, the best model is able to retrieve 95.9\% of the possible gold sentences after the document selection stage, suggesting that further improvements are more likely to come from document selection.

\begin{table}[h]
\small
    \centering
    \begin{tabular}{|l|l|l|l|}
    \hline
    Model & Acc. & Rec. & FEVER \\
    \hline
    Oracle & 84.2 & 94.7 & -- \\
    \hline
    \hline 
    Ptr & 74.6 & 86.1 & 68.6 \\
    \hline
    Our Baseline-RL & 74.6 & 87.5 & 69.2 \\
    \hline
    \hline 
    Our System & 76.74 & 90.84 & 73.17 \\
    \hline 
    \end{tabular}
    \caption{Ablation experiments on FV1-dev}
    \label{table:fever1_results}
\end{table}

Table \ref{table:fixers_fever2_results} evaluates the impact of the document pointer network and rule-based date handling on FV2-dev, as the impact of multi-hop reasoning and temporal relations is less visible on FV1-dev. We again compare Our Baseline-RL system to Our System and find an even larger 7.16 point improvement in FEVER score. We find that ablating the date post-processing (\textit{-dateProc}) and both the date post-processing and document ranking components (\textit{-dateProc,-docRank}) reduces the FEVER score by 1.45 and 3.5 points, respectively, with the latter largely resulting from a 5 point decrease in recall.

\begin{table}[h!]
\small
    \centering
    \begin{tabular}{|r|l|l|l|}
    \hline
    System & Acc. & Rec. & FEVER \\
    \hline
    \hline 
    Our System & 48.13 & 63.28 & 43.36 \\
    \hline 
        -dateProc & 45.14 & 63.28 & 41.91 \\
    \hline 
    -dateProc,-docRank & 44.29 & 58.32 & 39.86 \\
    \hline 
    \hline
    Our Baseline-RL & 44.04 & 57.56 & 36.2 \\
    \hline 
    \end{tabular}
    \caption{Ablation experiments on FV2-dev}
    \label{table:fixers_fever2_results}
\end{table}

\subsection{Ablation for Attack Types}
While Table \ref{table:breakers_data} presents the macro-average of all systems by attack type, we compare the performance of Our Baseline-RL and Our System in Table \ref{table:attack_type_results}.

Our System improves on evidence recall for multi-hop  claims (indicating that a multi-hop document retrieval step may help) and those with ambiguous entities or words (using a model to re-rank may remove false matches with high lexical similarity). For example, the claim \textit{``Honeymoon is a major-label record by Elizabeth Woolridge Grant.''} requires multi-hop reasoning over entities. Our System correctly retrieves the pages \textbf{[Lana Del Rey]} and \textbf{[Honeymoon (Lana Del Rey album)]}, but Our Baseline-RL is misled by the incorrect page \textbf{[Honeymoon]}. 
However, while recall increases on multi-hop
claims compared to the baseline, accuracy decreases, suggesting the model may be learning a bias of the claim or label distribution instead of relations between claims and evidence.

We also obtain large improvements on date manipulation examples (here a rule-based approach is better than our neural one); in contrast, multi-hop temporal reasoning leaves  room for improvement. For instance, for the claim \textit{``The MVP of the 1976 Canada Cup tournament was born before the tournament was first held,''} our full system correctly retrieves \textbf{[Bobby Orr]} and \textbf{[1976 Canada Cup]} (unlike the RL baseline). However, a further inference step is needed beyond our current capabilities -- reasoning that Orr's birth year (1948) is before the first year of the tournament (1976).

Finally, we enhance performance on multi-propositions as conjunctions or additional unverifiable information (indicating that relation sequence prediction helps). Claims (non-verifiable phrase in brackets) such as \textit{``Taran Killam is a [stage] actor.''} and \textit{``Home for the Holidays stars an actress [born in Georgia].''} are incorrectly predicted by the baseline even though correct evidence is retrieved.

\begin{table}
\small
    \centering
    \begin{tabular}{|l|l|l|l|l|}
    \hline
    Attack Type & & Acc. & Rec. & FEVER \\
    \hline
    \hline
    \multirow{2}{*}{Conjunction} & B & 16.95 & 92.0 & 16.95  \\
    \cline{2-5} 
     & S & \textbf{40.68}\textsuperscript{**} & 92.0 & \textbf{40.68}\textsuperscript{**} \\
    \hline 
    \hline 
    \multirow{2}{*}{Multi-hop} & B & 55.81\textsuperscript{*} & 29.07 & 19.77  \\
    \cline{2-5} 
     & S & 33.72 & \textbf{45.35}\textsuperscript{*} & 17.44 \\
    \hline
    \hline
    \multirow{2}{*}{Add. Unver.} & B & 48.0 & -- & 48.0  \\
    \cline{2-5}  
     & S & \textbf{80.0}\textsuperscript{**} & -- & \textbf{80.0}\textsuperscript{**} \\
    \hline
    \hline
    \multirow{2}{*}{Date Manip.} & B & 30.99 & 79.59 & 27.46 \\
    \cline{2-5} 
     & S & \textbf{53.52}\textsuperscript{***} & 79.59 & \textbf{42.25}\textsuperscript{**} \\
    \hline
    \hline
    \multirow{2}{*}{Multi-hop Temp.} & B & 3.33 & 10.34 & 0.0 \\
    \cline{2-5} 
     & S & 3.33 & \textbf{13.79} & 0.0 \\
    \hline
    \hline
    \multirow{2}{*}{Entity Disamb.} & B & 70.83 & 62.5 & 58.33 \\
    \cline{2-5} 
     & S & \textbf{79.17} & \textbf{79.17}\textsuperscript{*} & \textbf{70.83} \\
    \hline
    \hline
    \multirow{2}{*}{Lexical Sub.} & B & 33.33 & 65.71 & 25.0 \\
    \cline{2-5} 
     & S & 29.76 & \textbf{75.71}\textsuperscript{*} & \textbf{26.19} \\
    
    \hline 
    \end{tabular}
    \caption{Attack results for our FV2-dev claims. \textbf{B:} Our Baseline-RL, \textbf{S:} Our System. \textbf{*}: $p<0.05$ \textbf{**}: $p<0.01$  \textbf{***}: $p<0.001$ by approximate randomization test}
    \label{table:attack_type_results}
\end{table}

\section{Conclusion}
We showed weaknesses in approaches to fact-checking via novel adversarial claims.  We took steps towards realistic fact-checking with targeted improvements to multi-hop reasoning (by a document pointer network and a pointer network for sequential joint sentence selection and relation prediction), simple temporal reasoning (by rule-based date handling), and ambiguity and variation (by fine-tuned contextualized  representations).

There are many unaddressed vulnerabilities that are relevant for fact-checking. The Facebook bAbI tasks \cite{Weston2016TowardsAQ} include other types of reasoning (e.g. positional or size-based). The DROP dataset \cite{dua-etal-2019-drop} requires mathematical operations for question answering such as addition or counting. Propositions with causal relations \cite{hidey-mckeown-2016-identifying}, which are event-based rather than attribute-based as in FEVER, are also challenging. Finally, many  verifiable claims are non-experiential \cite{park-cardie-2014-identifying}, e.g. personal testimonies, which would require predicting whether a reported event was actually possible. 

Finally, our system could be improved in many ways. Future work in multi-hop reasoning could  represent the relation between consecutive pieces of evidence and future work in temporal reasoning could incorporate numerical operations with BERT \cite{andor-etal-2019-giving}.
One limitation of our system is the pipeline nature, which may require addressing each type of attack individually as adversaries adjust their techniques.
An end-to-end approach or a query reformulation step (re-writing claims to be similar to FEVER) might make the model more resilient as new  attacks are introduced.

\section*{Acknowledgements}
The authors thank Kathy McKeown, Chris Kedzie, Fei-Tzin Lee, and Emily Allaway for their helpful comments on the initial draft of this paper and the anonymous reviewers for insightful feedback.

\bibliography{acl2020}
\bibliographystyle{acl_natbib}

\appendix

\section{Examples of Attack Types}
\label{appendix:attack_examples}
Table \ref{table:fever_summary} displays examples for each type of attack. The multi-propositional examples include attacks for {\sc conjunction}, {\sc multi-hop reasoning}, and {\sc additional unverifiable propositions}. For temporal reasoning, we provide examples for {\sc date manipulation} and {\sc multi-hop temporal reasoning}. The lexical variation examples consist of {\sc entity disambiguation} and {\sc lexical substitution}.

\begin{table*}
\footnotesize
\centering
\begin{tabular}{|p{2cm}|p{3.5cm}|l|p{8cm}|}
     \hline
    Attack Type & Example Claim & Label & Evidence \\
     \hline
      Conjunction & Blue Jasmine has Sally Hawkins acting in it \textbf{and} \st{Blue Jasmine} was filmed in San Francisco. & NEI & N/A \\
     \hline
     Multi-Hop Reasoning & Goosebumps was directed by \st{Rob Letterman} \textbf{the person who co-wrote Shark Tale}. & S & \textbf{[Goosebumps (film)]} It was directed by Rob Letterman, and written by Darren Lemke, based from a story by Scott Alexander and Larry Karaszewski. \textbf{[Rob Letterman]} Before Letterman's film subjects took him into outer space with Monsters vs. Aliens (2009), he was taken underwater, having co-directed and co-written Shark Tale. \\
     \hline
     Additional Unverifiable Propositions & Roswell is an American TV series \textbf{with 61 episodes}. & NEI & N/A\\
     \hline
     \hline
     Date Manipulation& Artpop was Gaga's second consecutive number-one record in the United States \st{in 2009} \textbf{before 2010}. & {\sc r} & \textbf{[Artpop]} Gaga began planning the project in 2011, shortly after the launch of her second studio album, Born This Way.\\
     \hline
     Multi-Hop Temporal Reasoning & Lisa Murkowski's father resigned from the Senate after serving as Senator. & {\sc s} & \textbf{[Lisa Murkowski]} She is the daughter of former U.S. Senator and Governor of Alaska Frank Murkowski. Murkowski was appointed to the U.S. Senate by her father, Frank Murkowski, who resigned his seat in December 2002 to become the Governor of Alaska. \textbf{[Frank Murkowski]} He was a United States Senator from Alaska from 1981 until 2002 and the eighth Governor of Alaska from 2002 until 2006. \\
     \hline
     \hline
     Entity Disambiguation & Kate Hudson is a left wing political activist & S & \textbf{[Kate Hudson (activist)]} Katharine Jane ``Kate'' Hudson (born 1958) is a British left wing political activist and academic who is the General Secretary of the Campaign for Nuclear Disarmament (CND) and National Secretary of Left Unity.\\
     \hline
     Lexical Substitution & The Last Song began \st{filming} \textbf{shooting} on Monday June 14th 2009. & {\sc r} & \textbf{[The Last Song (film)]} Filming lasted from June 15 to August 18, 2009 with much of it occurring on the island\'s beach and pier.\\
     \hline
\end{tabular}
\caption{Examples of the seven sub-types of attacks.  Claims edited with word substitution or insertion have their changes in bold. Deletions are marked in strikethrough. Wikipedia titles are represented in bold with square brackets. \textbf{S:} {\sc SUPPORTS} \textbf{R:} {\sc REFUTES} \textbf{NEI:} {\sc NOT ENOUGH INFORMATION} }
\label{table:fever_summary}
\end{table*}

\section{Hyper-parameters and Experimental Settings}
\label{appendix:hyps}

We select $M=30$ Wikipedia articles using TF-IDF when combining with our other candidate document selection methods and select $D=5$ after document ranking.  We select $N=5$ sentences during sentence selection, consistent with the shared task evaluation.

\subsection{BERT Language Model Fine-Tuning}
We use version 0.5.0 of the Huggingface library (\url{https://github.com/huggingface/pytorch-pretrained-BERT}) to fine-tune the ``BERT-base'' model using the default settings. We lowercase all tokens and use the default BERT tokenizer.

\paragraph{Document Ranking} Our dataset of title and claim pairs (obtained from FV1-train) consists of 140,085 positive examples and 630,265 negative examples in training with approximately 10\% set aside for validation (16,016 positive examples and 84,437 negative).  As recommended by \newcite{bert}, we select hyper-parameters by grid search over 16 and 32 for batch size, 2e-5, 3e-5, and 5e-5 for learning rate, and 3 and 4 for the number of epochs.

\paragraph{Sentence Selection} Our dataset of sentence and claim pairs (also obtained from FV1-train) consists of 54,431 {\sc s} relations, 54,592 {\sc r} relations, and 54,501 {\sc nei} relations in training, with approximately 10\% set aside for validation (6,139 {\sc s} relations, 5,984 {\sc r} relations, and 6,050 {\sc nei} relations).
We again select hyper-parameters consistent with the recommended best practice.

\subsection{Pointer Network} We train both the document ranking and sentence selection pointer networks on FV1-train with the same hyper-parameters using Adagrad \cite{duchi2011adaptive} with a learning rate of 0.01, a batch size of 16, and a maximum of 140 epochs with early stopping on FV1-dev. The dimension of the pointer network LSTM hidden state is set to 200 with $q=3$ hops over the memory. We use a beam width of 5 during inference. The MLP used to predict relations has a hidden layer dimensionality of 200 and we set $\lambda=1$.

\subsection{Reinforcement Learning}

To make the sentence extractor an RL agent, we can formulate a Markov Decision Process (MDP): at each
extraction step $t$, given a claim $c$, the agent observes the current
state and samples an action from Equation 3 to extract a document sentence $s$, predict the relation label $l$ and receive a reward $r(t + 1) = \text{FEVER}(c, s, l)$. 
We train using REINFORCE, adapted with an Actor-Critic to minimize variance (detailed by \newcite{P18-1063}). As RL often requires pre-training, we combine the pointer network loss from Equation 3 with the RL loss ($\mathcal{L}(\theta_{rl})$) and the relation prediction loss ($\mathcal{L}(\theta_{rel}$):
\begin{equation}
    \mathcal{L}(\theta) = \lambda_1 \mathcal{L}(\theta_{ptr}) + \lambda_2 \mathcal{L}(\theta_{rl}) + \mathcal{L}(\theta_{rel})
    \label{equation:builders}
\end{equation}
We set both $\lambda_1=1$ and $\lambda_2=1$.

\end{document}